# Jointly Learning Clinical Entities and Relations with Contextual Language Models and Explicit Context

Paul Barry[1], Sam Henry[1], PhD, Meliha Yetisgen[2], PhD, Bridget McInnes[3], PhD, Özlem Uzuner[1], PhD, FACMI
[1]George Mason University, Fairfax, VA, [2]University of Washington, Seattle, WA, [3]Virginia Commonwealth University, Richmond, VA

**Introduction**

Electronic health records contain a wealth of information about a patient's health. However, much of this information is stored as unstructured narrative text. Accessing this information requires natural language processing (NLP) techniques that can extract knowledge from the text. Named Entity Recognition (NER) and Relation Extraction (RE) automatically identify mentions of salient entities and the relationships between them in text, making them accessible to computerized systems. This makes NER and RE vital components of many applications, such as information retrieval, question answering, and clinical decision support. NER and RE are traditionally performed sequentially, where entities are first identified through NER and the relations between the predicted entities are inferred by RE. Recently, deep learning systems have shown great success in both tasks. However, they have drawbacks when applied sequentially. (1) Sequential application does not allow the two tasks to inform each other. (2) It is computationally wasteful since NER and RE duplicate some of their processing. (3) Errors in NER directly limit performance in RE. Multi-task Learning (MTL) architectures address these limitations (1) by allowing the two tasks to be learned jointly (i.e., simultaneously) and to inform each other, (2) by eliminating redundant work, and (3) by eliminating cascading errors. MTL approaches have shown great success for joint NER/RE in the clinical domain.[1,2,3] Past joint NER/RE methods use contextual language models to build representations of relations using only the entities and do not explicitly include their context. This results in significant reliance on the language models to encode relation information.

We hypothesize that explicit integration of contextual information into an MTL framework would emphasize the significance of context for boosting performance in joint NER/RE. We create an MTL architecture that uses a novel average pooling operation on the word embeddings before, between, and after both entities, in addition to both entities, providing additional context to the RE classifier. Combining this idea with MTL creates a model capable of accurately predicting both named entities and the relationships between them, all while using additional context for each relationship inference.

**Methods**

Inspired by the context representation of Luo et al.[4], we segment sentences into the two entities along with the words before, between, and after those entities in order to represent context. Different from Luo et al.[4], we pool on non-overlapping segments creating vector representations which explicitly include the surrounding context. Soares et al.[5] and Wu & He[6] introduce the use of pooling operations on the embeddings of entities for use in relation extraction. We extend this idea beyond only the entities to include the surrounding words by applying average pooling to the segments before, between, and after both entities. Additionally, we assume a single sentence can contain many relationships and thus train the system to produce meaningful relationship representations without reliance on special tokens to mark the entities since doing so can exponentially increase compute time. We integrate this context representation and the RE system with a bi-directional Recurrent Neural Network (RNN) for NER to form our MTL architecture. We evaluate the architecture on the 2010 i2b2/VA dataset[7], achieving promising results.

In Figure 1, we show that our system takes a single sentence as input (1) and outputs the named entities (6) and relations (9) appearing within the sentence. Without preprocessing, our system uses BERT[8] (2) to build context specific representations of each word (3). Embeddings generated by BERT are passed into a three layer, bidirectional RNN (4) with a Tanh activation function then to a FFNN (5) with a ReLU activation to infer the entity type (6). When training, BERT embeddings and entity boundaries are input to the RE system. Using the entity boundaries, the RE system novelly segments the sentence (7) into five pieces then conducts an average pooling operation (7) on each segment to generate embeddings of a consistent size. These embeddings are concatenated to create a single relation representation which is passed through a FFNN (8) with a ReLU activation function to infer the relation type (9). Adam optimizer and Cross-Entropy Loss are used for both NER and RE. Back-propagation is done using the summed error of both tasks. Argmax is used to select the predicted class for both NER and RE.

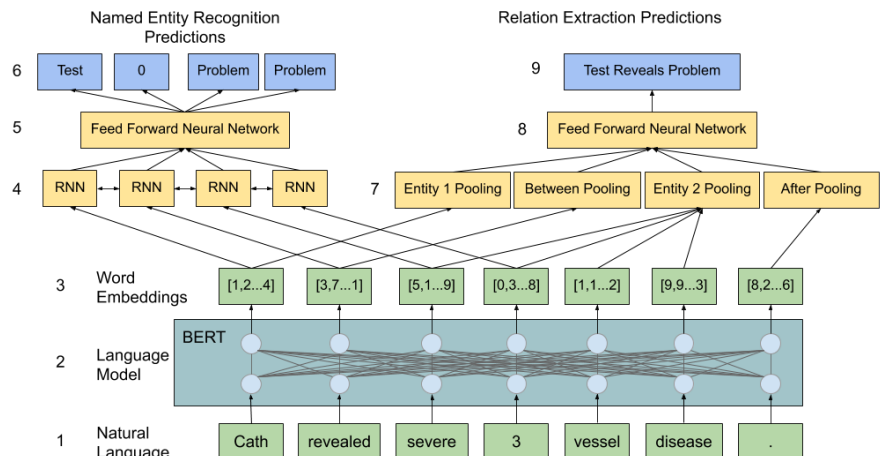

Figure 1: Diagram of our jointly learned NER and RE system. The BERT language model is a 12 layer, self-attention neural network (2). NER system uses a 3 layer RNN (4) followed by a single layer FFNN (5). RE system uses average pooling (7) and a single layer FFNN (8).

**Results**

We utilized the 2010 i2b2/VA concept and relation classification challenge dataset[7] for learning entities and relations. The dataset is composed of clinical notes annotated for mentions of problems, tests, and treatments along with their relationships. We used 80% of the training data for development and training, 20% for validation. Reported results use the held-out test set of 2010 i2b2/VA challenge. When training, we used a batch size of 32 due to memory limitations, learning rate of 1e-5, and early stopping threshold of 10 epochs. Downsampling was done by entity pair combination such that positive and negative problem-to-problem relations were 1:4, test-to-problem was 1:2, and treatment-to-problem was 1:1.

Using the above architecture we created NER and RE classifiers with performance close to the state of the art (SOTA). The 2010 i2b2/VA RE task[7] used micro-averaged F1 score for the RE and NER tasks, and exact entity matching for NER. SOTA performance for RE is 73.7 F1[9] while ours achieved a 72.97 F1. For NER, the single model SOTA performance is 86.84 F1[10] while ours yields 85.43 F1. The joint NER/RE performance is measured by the RE system's F1 score when using the NER classifier's predicted entities. We achieved a 49.07 F1 score while the SOTA system achieved a 48.4 F1[9].

**Discussion**

We developed a novel MTL approach to address problems inherent to sequential NER and RE systems by jointly learning both tasks in unison using a method to explicitly incorporate context into the RE classifier. Future work includes experimentation with architectures that utilize attention mechanisms in the RE classifier and measuring the architecture's performance on a broader spectrum of clinical datasets.